\documentclass[10pt,twocolumn,letterpaper]{article}

\usepackage{wacv}
\usepackage{times}
\usepackage{epsfig}
\usepackage{graphicx}
\usepackage{amsmath}
\usepackage{amssymb}
\usepackage[accsupp]{axessibility}  % Improves PDF readability for those with disabilities.

\usepackage{enumitem}
\usepackage{multirow}
\usepackage{xcolor}
\usepackage{adjustbox}
\newcommand{\widthscale}{0.125}
 
\usepackage[toc,page]{appendix}
\usepackage{caption}
\usepackage{comment}
\usepackage{subcaption}
\usepackage{bbding}
\newcommand{\V}{\mathcal{V}}
\newcommand{\E}{\mathcal{E}}
\newcommand{\G}{\mathcal{G}}
\newcommand{\N}{\mathcal{N}}
\newcommand{\R}{\mathbb{R}}

\renewcommand{\paragraph}[1]{\vspace{0.2cm}\noindent\textbf{#1}~~}

\def\wacvPaperID{501} % Enter the WACV Paper ID here

\wacvfinalcopy % *** Uncomment this line for the final submission

\ifwacvfinal
\fi

\ifwacvfinal
\usepackage[breaklinks=true,bookmarks=false]{hyperref}
\else
\usepackage[pagebackref=true,breaklinks=true,colorlinks,bookmarks=false]{hyperref}
\fi

\begin{document}

%%%%%%%%% TITLE
%%%%%%%%% TITLE
% \title{\LaTeX\ Author Guidelines for WACV Proceedings}
\title{MEGAN: Memory Enhanced Graph Attention Network for Space-Time Video Super-Resolution}

\author{Chenyu You$^{1}$ \quad Lianyi Han$^{2}$ \quad Aosong Feng$^1$ \quad Ruihan Zhao$^3$ \quad \\
Hui Tang$^2$\quad Wei Fan$^2$\\
$^1$Yale University \quad
$^2$Tencent Hippocrates Research Lab \\
$^3$The University of Texas at Austin}

% \title{\LaTeX\ Author Guidelines for WACV Proceedings}

% \author{First Author\\
% Institution1\\
% Institution1 address\\
% {\tt\small firstauthor@i1.org}
% % For a paper whose authors are all at the same institution,
% % omit the following lines up until the closing ``}''.
% % Additional authors and addresses can be added with ``\and'',
% % just like the second author.
% % To save space, use either the email address or home page, not both
% \and
% Second Author\\
% Institution2\\
% First line of institution2 address\\
% {\tt\small secondauthor@i2.org}
% }

\maketitle

\ifwacvfinal
\thispagestyle{empty}
\fi

\begin{abstract}

  Space-time video super-resolution (STVSR) aims to construct a high space-time resolution video sequence from the corresponding low-frame-rate, low-resolution video sequence. Inspired by the recent success to consider spatial-temporal information for space-time super-resolution, our main goal in this work is to take full considerations of spatial and temporal correlations within the video sequences of fast dynamic events. To this end, we propose a novel one-stage memory enhanced graph attention network (MEGAN) for space-time video super-resolution. Specifically, we build a novel long-range memory graph aggregation (LMGA) module to dynamically capture correlations along the channel dimensions of the feature maps and adaptively aggregate channel features to enhance the feature representations. We introduce a non-local residual block, which enables each channel-wise feature to attend global spatial hierarchical features. In addition, we adopt a progressive fusion module to further enhance the representation ability by extensively exploiting spatial-temporal correlations from multiple frames. Experiment results demonstrate that our method achieves better results compared with the state-of-the-art methods quantitatively and visually.

\end{abstract}

\section{Introduction}

Space-time video super-resolution (STVSR) \cite{shechtman2005space} aims at reconstructing a photo-realistic video sequence with high spatial and temporal resolutions from its corresponding low-frame-rate (LFR) and low-resolution (LR) counterparts. High-resolution (HR) slow-motion video sequences provide more visually appealing details in the space-time domain, which finds a wide range of applications, such as sports video, movie making, and medical imaging analysis.

% \begin{figure}[t]
% \centering
% \includegraphics[width=0.9\columnwidth]{latex/Figs/long-term-memory.pdf} % Reduce the figure size so that it is slightly narrower than the column. Don't use precise values for figure width.This setup will avoid overfull boxes.
% \caption{Overview of Long-Term Memory (LMGA) Block. We randomly select~$\tau$ frame features from the shuffled ordered-index frame feature maps~$\{1,\ldots,2n+1\}$, and then cache the selected features to form the global memory block. By leveraging these cached information, the global-local feature aggregation function $N_g{(\cdot)}$ enables the key frame to access more complex spatial-temporal information.}
% \label{fig:long-term-memory}
% \end{figure}

%%%%%%%%%%%%%%%%%%%%%%%%%
\begin{figure*}
    \centering
    \includegraphics[width=0.95\textwidth]{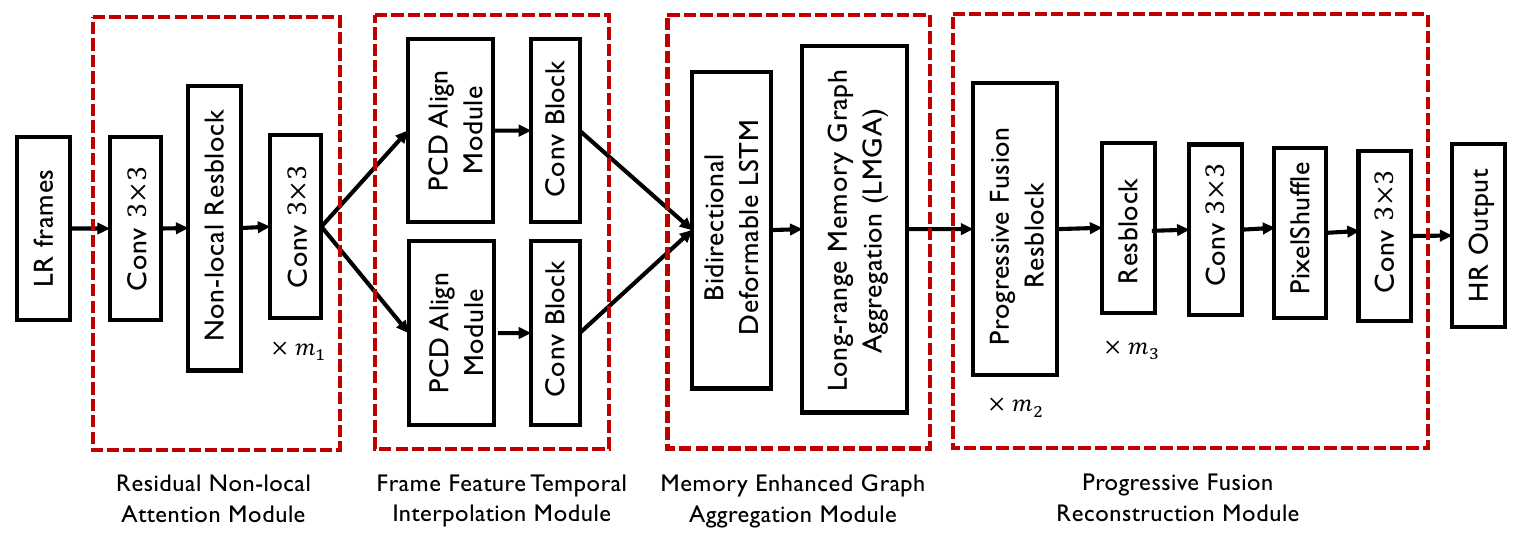}
    \vspace{-3mm}
    \caption{Framework overview of Memory Enhanced Graph Attention Networks (MEGAN). Our approach uses the residual non-local attention module to extensively exploit information both in temporal and spatial dimensions. Then we apply frame feature temporal interpolation and long-range memory graph aggregation (LMGA) module to handle complex motions in a dynamic scene. By capturing global temporal features, it enables progressive function reconstruction to generate HR output.}
    \label{fig:megan}
    \vspace{-3mm}
\end{figure*}

In recent years, there is a surge of research interests in upscaling the spatial-temporal resolution of a video simultaneously. Most of traditional space-time super-resolution (SR) methods~\cite{shahar2011space,faramarzi2012space,li2015space,mudenagudi2010space} propose to increase the resolution both in the time and space domain by combining information from multiple low-resolution video recordings in a dynamic scene, where they address the space-time SR problem as an optimization problem. For example, Shechtman~\etal~\cite{shahar2011space} formulates the problem of the space-time SR as linear least-square minimization with suitable regularization priors. Specifically, they demonstrate that it is possible to obtain an improved HR video sequence in both spatial and temporal dimensions. However, the complex nature of the real-world video sequences limits the effectiveness of these methods since they can only super-resolve video sequences containing slow-varying motions with a simple analytic form. Furthermore, the high computational cost limits their practical use in real-world scenarios.

Deep convolutional neural networks (CNNs) have shown promising results in various machine learning tasks \cite{su2020audeo,you2020unsupervised,you2021momentum,kong2020sia,kong2019adaptive,you2021simcvd,you2020data,you2021knowledge,you2021sbilsan,xu2021semantic,cheng2016hybrid,you2018structurally,you2019ct,you2019low,li2021assessing,lyu2018super,li2019novel,cheng2016identification,cheng2016random,cheng2017body,chen2021self,you2021self,tang2021recurrent,tang2021spatial,yang2020nuset,guha2020deep,tang2019clinically,tang2019nodulenet,you2020contextualized,you2021mrd,liu2021auto}, such as video frame interpolation (VFI)~\cite{Niklaus_ICCV_2017,xue2019video,jiang2018super}, video super-resolution (VSR)~\cite{wang2019edvr,RBPN2019,tian2018tdan}, and image recognition~\cite{he2016deep,tang2019end,tang2018automated}. To jointly up-scaling videos both in space and time, the most common way is to perform learning-based VFI and VSR methods alternately and independently. For example, it first constructs in-between frames by VFI, and then generates high-resolution video sequences by VSR. However, due to the strong space-time relationship among video sequences, such two-stage methods may fail to capture correlations between space and time, which is essential for increasing spatial-temporal resolution. Moreover, these methods consist of two independent CNN-based networks, which are computationally expensive.
% and inefficient in practice

To enhance feature representations used for space-time video SR purposes, it is desirable to integrate the spatial and temporal information. Existing one-stage space-time SR approaches enhance feature representations by jointly aggregating the representations both in spatial and temporal dimensions. However, these methods have two major limitations. First, these learned feature representations only contain local contexts, while lacking global structural information. Features from multiple frames and different locations can help the network to model long-term dependencies among the video sequences more efficiently. Thus, only considering spatial information is an insufficient approximation of the global and local influence. Second, space-time SR is a dynamic process with complex scenes. Most CNN-based methods indistinguishably extract different types of features to super-solve in space-time domain, which lack discriminative representation ability to attend to different types of information of fast dynamic events.

\begin{figure*}[t]
\centering
\includegraphics[width=0.9\textwidth]{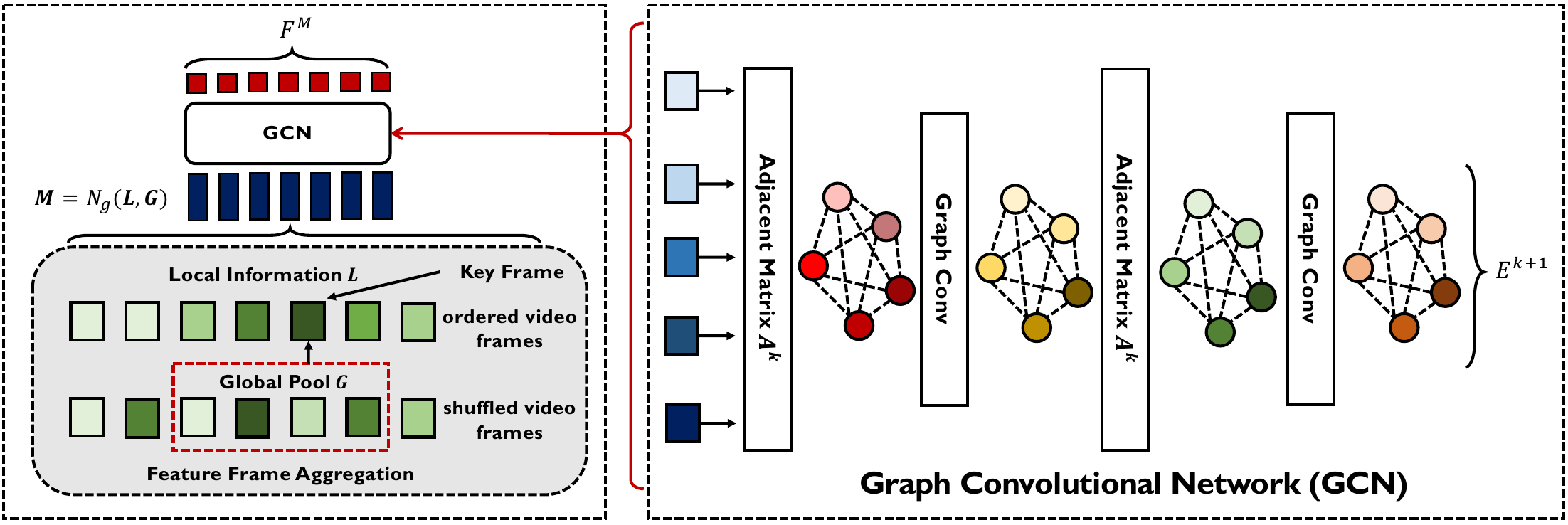} % Reduce the figure size so that it is slightly narrower than the column. Don't use precise values for figure width.This setup will avoid overfull boxes.
\vspace{-3mm} 
\caption{Overview of Long-range Memory Graph Aggregation (LMGA) module. The LMGA includes two operations: frame feature aggregation and graph construction. We randomly select~$\tau$ frame features from the shuffled ordered-index frame feature maps~$\{1,\ldots,2n+1\}$, and then cache the selected features to form the global memory block. By leveraging these cached information, the global-local feature aggregation function $N_g{(\cdot)}$ enables the key frame to access more complex spatial-temporal information. We also dynamically construct $\G_{k}$ by Graph Construction to exploit the spatial-temporal relationship between input feature frames in a discriminative manner. We then construct the output features by the small network containing two convolutional layers.}
\label{fig:long-term-memory}
\vspace{-3mm} 
\end{figure*}

In this paper, we work toward simultaneously reconstructing a video sequence of high spatial and temporal resolution given a low-frame-rate, low-resolution video sequence. The objective is achieved by the proposed memory enhanced graph attention network (MEGAN), which comprehensively leverages the spatial-temporal information to enhance the feature representation capability of CNNs. The motivation comes from the fact that features in spatial domain correlate with the contexts in temporal domain. In implementation, we use the non-local residual block to capture the global dependencies between channel-wise features and encourage the model to capture more informative features in the spatial dimension. Considering the non-local block requires high computational cost, which is practically challenging to use for the task of STVSR. Therefore, we improve the non-local residual block to boost convergence. In addition, we design a novel long-range memory graph aggregation (LMGA) module to adaptively learn a robust spatial-temporal feature representation between the current key frame and the neighboring frames. Specifically, we cache the extracted hierarchical features in LMGA and combine embedded features in the collection to propagate contextual information towards the improved feature representations by constructing the adaptive graph. The embedded spatial features and temporal contexts provide a more clear definition of the feature aspects. To further promote the capability of feature representation, we apply progressive fusion residual dense blocks~(PFRDBs) to progressively extract both intra-frame spatial correlations and inter-frame temporal correlations. Experimental results show that the proposed method achieves better results compared with the state-of-the-art methods.
Our contributions are summarized as follows:
\begin{itemize}
    \item We present a unified graph attention network for space-time SR problems. Our method utilizes the non-local residual blocks to better learn different kinds of information (\eg, low-and high-frequency information) and long-range spatial dependencies.
    \vspace{-1mm}
    \item We devise a novel long-range memory graph aggregation (LMGA) module which enables the key frames to adaptively model temporal dependency among the video sequences of fast dynamic events. Both local and global information of the rapid dynamic space-time scene can allow the network to obtain an improved video sequence of high spatial-temporal resolution.
    \vspace{-1mm} 
    % To the best of our knowledge, this is the first time to consider graph convolutional networks (GCNs) for video enhancement problems.
    \item We utilize progressive fusion residual dense blocks (PFBDNs) to combine spatial features and temporal information for better reconstruction performance.
    \vspace{-1mm}
    \item We validate the effectiveness of our MEGAN with extensive experiments. MEGAN achieves superior performance compared with state-of-the-art methods on three benchmark datasets.
    % \vspace{-3mm}
\end{itemize}

% \begin{figure}[t]
% \centering
% \includegraphics[width=0.9\columnwidth]{latex/Figs/non-local.pdf} % Reduce the figure size so that it is slightly narrower than the column. Don't use precise values for figure width.This setup will avoid overfull boxes.
% \caption{Non-local Residual Block. We illustrate feature map as their shape in term of ~$H\times W\times C$. Red dashed line denotes matrix reshaping. $\otimes$ refers to matrix multiplication. $\oplus$ refers to element-wise addition.}
% \label{fig:non-local}
% \end{figure}

%-------------------------------------------------------------------------
\section{Related Work}
% Super resolution has been a widely studied problem in computer vision and still remains a open question in recent investigations of deep learning. 
In this section, we overview the progress on three related topics: video super-resolution, video frame interpolation, and space-time video super-resolution.

% \begin{description}[style=unboxed,leftmargin=0cm]

\paragraph{Video Super-Resolution}
Video super-resolution aims to recover HR video frames from the corresponding corrupted video frames. Several works~\cite{caballero2017real,xue2019video,tao2017detail,sajjadi2018frame} adopted optical flow for explicit temporal alignment in video sequences, which can effectively capture image details among multiple frames and inter-frame motion to improve reconstruction accuracy. Other works~\cite{kim2016accurate,tai2017image,zhang2018residual} utilized deep residual learning to reconstruct HR output. Some works~\cite{jo2018deep,tian2018tdan,wang2019edvr} attempted to address the implicit temporal alignment without motion estimation between video sequences at subpixel and subframe accuracy.~For example, Jo~\etal~\cite{jo2018deep} generated dynamic upsampling filters and residuals to recover HR frames by preserving temporal consistency.~Tian~\etal~\cite{tian2018tdan} and Wang~\etal~\cite{wang2019edvr} adaptively learned the temporal information without computing optical flow. Considering that recurrent neural networks (RNNs) can model long-term temporal information, some recent efforts~\cite{huang2015bidirectional,sajjadi2018frame,haris2019recurrent} have been made to adopt RNN-based networks to encourage temporal dependency modeling without explicit temporal alignment, and therefore can steer very complex scene dynamics.

% RNN-based networks~\cite{huang2015bidirectional,sajjadi2018frame,haris2019recurrent} haven been proposed to
% BRCN~\cite{huang2015bidirectional}, FRVSR~\cite{sajjadi2018frame}, and RBPN~\cite{haris2019recurrent} proposed RNN-based networks to encourage temporal dependency modeling without explicit temporal alignment, and therefore can steer very complex scene dynamics.

\paragraph{Video Frame Interpolation}
Video frame interpolation aims at synthesizing the in-between frames from two consecutive video frames. Video frame interpolation approach can usually be categorized as kernel-based~\cite{DAIN,Niklaus_ICCV_2017,Niklaus_CVPR_2017}, phase-based~\cite{meyer2015phase,meyer2018phasenet}, and flow-based~\cite{niklaus2018context,liu2017video,jiang2018super,xue2019video}. Some recent kernel-based works~\cite{Niklaus_ICCV_2017,Niklaus_CVPR_2017} employed CNNs to estimate adaptive filters for each pixel and then convolved with input frames to synthesize an intermediate frame. More recently, some efforts~\cite{meyer2015phase,meyer2018phasenet} proposed to synthesize an intermediate frame by utilizing a neural network decoder to estimate a per-pixel phase-based motion representation. Most recently, the common approach is to guide the synthesis of an intermediate frame by predicting optical flow between two input frames. For example, Context-Aware Synthesis (CtxSyn)~\cite{niklaus2018context} computed the optical flow between two input frames and then forward-warped the images with temporal contextual information extracted from input images corresponding to the optical flow for high-quality frame synthesis.

\paragraph{Space-Time Video Super-Resolution}
The pioneer space-time super-resolution (SR) work~\cite{shechtman2005space} addressed the problem of simultaneous spatial and temporal super-resolution by solving linear function. The following work~\cite{shahar2011space} achieved space-time SR from a single video by combing information from multiple space-time patches at sub-frame accuracy. More recently, the success of CNNs has motivated the development of STVSR methods~\cite{xiang2020zooming,haris2020space,kim2019fisr,kang2020deep,xiao2020space,xiao2021space,tseng2021dual}. Xiang~\etal~\cite{xiang2020zooming} proposed a one-stage STVSR method which employed a deformable convolutional LSTMs (ConvLSTM) to align and aggregate temporal contexts for better utilization of global information. Haris~\etal~\cite{haris2020space} introduced a method to super-solve in spatial and temporal dimensions simultaneously and jointly. However, most methods have limited capacity to capture global dependencies and handle space-time visual patterns in complex dynamic scenes.

% \end{description}
%-------------------------------------------------------------------------
\section{Method}
Given a LR space-time video sequence~(${I}^{LR} = \{I^{LR}_{2t-1}\}^{n+1}_{t=1}$), our method aims at reconstructing the corresponding HR video sequence~(${I}^{HR} = \{I^{HR}_{t}\}^{2n+1}_{t=1}$). The size of input LR video frame is $H\times W$, where $H$ and~$W$ refer to height and width, respectively. The size of HR output is $rH\times rW$ with the spatial upscaling factor~$r$. An overview of MEGAN for space-time video super-resolution is shown in Figure~\ref{fig:megan}. The framework consists of four main parts: residual non-local attention module, frame feature temporal interpolation module, memory enhanced graph aggregation module, and progressive fusion reconstruction module.

\subsection{Residual Non-local Attention Module}
Inspired by \cite{liu2017robust,liu2018non,wang2018non,yi2019progressive}, we first use a $3\times 3$~convolutional (Conv) layer as a shallow feature extractor. Then, we incorporate a residual non-local attention block~\cite{wang2018non,yi2019progressive} to capture long-range dependencies between channel-wise features and allow the model to discriminatively learn different types of information (e.g., low and high frequency information). Mathematically, the non-local operation is defined as follows:
\begin{equation}
    y_{i} =  \left( \sum_{\forall j} f(x_i,x_j)g(x_j)\right) / \sum_{\forall j} f(x_i,x_j),
	\label{eq: non_local}
\end{equation}
where $x$ denotes the input of non-local operation and $y$ denotes the output with the same size as $x$. $i$ is the output position index, and $j$ is the index of all possible positions. The pairwise function $f(\cdot)$ computes the attention between two inputs, while the function $g(\cdot)$ computes the feature representation at a certain position. Non-local neural networks~\cite{wang2018non} provide several versions of $f(\cdot)$, such as Gaussian function, dot product similarity, and feature concatenation. Different from \cite{yi2019progressive}, we use the dot product similarity to evaluate the pairwise attention:
% As illustrated in Figure~\ref{fig:non-local}, we consider the dot product similarity to evaluate the pairwise attention:
\begin{equation}
\vspace{-1mm}
    f(x_i,x_j)=u(x_i)v(x_j)=(W_{u}x_i)W_{v}x_j,
	\label{eq: emb_gaussian}
\vspace{-0.5mm}
\end{equation}
where $W_{u}$ and $W_{v}$ are weight matrices, and we adopt a linear embedding for $g: g(x_{j})=W_{g}x_{j}$ with the weigtht matrix $W_{g}$. The reason why we use dot product similarity rather than the Gaussian function is to boost convergence, which allows us to train very deep networks, being more suitable for STVSR. Thus, we have the output $z$ at position $i$ of the non-local residual block:
\begin{multline}
\vspace{-3mm}
    z_{i}\!=\!W_{z}y_{i}+x_{i}\!=\!W_{z}\text{softmax}(f(x_i,x_j))g(x_j) + x_{i},
	\label{eq: sofmax}
\end{multline}
where $W_{z}$ is a weight matrix. After the $m_{1}$ Conv layers, the hierarchical feature maps~(${F}^{LR} = \{F^{LR}_{2t-1}\}^{n+1}_{t=1}$) serve as inputs for the following frame feature interpolation module.

\begin{table*}[t]
% Quantitative comparison of our results and two-stage VFI and VSR methods on testsets. The best two results are highlighted in \textcolor{red}{red} and \textcolor{blue}{blue} colors, respectively. The total runtime is measured on the entire Vid4 dataset \cite{liu2011bayesian}. Note that we omit the baseline models with Bicubic when comparing in terms of runtime.}
\centering
\resizebox{\textwidth}{!}{
\begin{tabular}{c c|c c c c c c c c c c c c}
\hline\hline
VFI & VSR & \multicolumn{2}{c}{Vid4} & \multicolumn{2}{c}{Vimeo-Fast} & \multicolumn{2}{c}{Vimeo-Medium} & \multicolumn{2}{c}{Vimeo-Slow} & \multicolumn{2}{c}{Adobe240} & \multirow{2}{*}{\begin{tabular}[c]{@{}c@{}}Parameters\\   (Million)\end{tabular}}\\
Method & Method & PSNR & SSIM & PSNR & SSIM & PSNR & SSIM & PSNR & SSIM & PSNR & SSIM \\
\hline\hline
SuperSloMo~\cite{jiang2018super} & Bicubic & 23.38 & 0.6154     & 34.94 & 0.8848        & 30.35 & 0.8570        & 28.82 & 0.8230 & 26.34 & 0.7361 & 19.8 \\
SuperSloMo~\cite{jiang2018super} & RCAN~\cite{zhang2018residual} & 23.81 & 0.6401 & 34.54 & 0.9076			& 32.50 & 0.8884 			& 30.69 & 0.8624 & 26.43 & 0.7398 & 19.8+16.0 \\
SuperSloMo~\cite{jiang2018super} & RBPN~\cite{haris2019recurrent} & 24.42 & 0.7105      & 34.94 & 0.9174 			& 33.23 & 0.9059 			& 31.04 & 0.8753  & 28.22 & 0.8170 & 19.8+12.7 \\
SuperSloMo~\cite{jiang2018super} & EDVR~\cite{wang2019edvr} & 25.33 & 0.7417    & 35.19 & 0.9201    & 33.51 & 0.9096     & 31.42 & 0.8832       & 28.72 & 0.8320 & 19.8+20.7 \\
% SuperSloMo~\cite{jiang2018super} & PFRB~\cite{yi2019progressive} & - & - 										& - & - 			& - & - 			& - & - & - & - \\
\hline
SepCov~\cite{Niklaus_ICCV_2017} & Bicubic & 23.62 & 0.6322      & 32.42 & 0.8901        & 30.77 & 0.8650      & 29.18 & 0.8310        & 26.61 & 0.7457    & 21.7\\
SepCov~\cite{Niklaus_ICCV_2017} & RCAN~\cite{zhang2018residual} & 24.92 & 0.7236    & 34.97 & 0.9195 			& 33.59 & 0.9125    & 32.13 & 0.8967    & 26.84 & 0.7488 & 21.7+16.0\\
SepCov~\cite{Niklaus_ICCV_2017} & RBPN~\cite{haris2019recurrent} & 25.63 & 0.7546       & 35.21 & 0.9244      & 34.19 & 0.9231        & 32.86 & 0.9090        & 30.18 & 0.8706 & 21.7+12.7\\
SepCov~\cite{Niklaus_ICCV_2017} & EDVR~\cite{wang2019edvr} & 25.90 & 0.7785     & 35.24 & 0.9252     & 34.23 & 0.9240        & 32.97 & 0.9112    & 30.28 & 0.8745 & 21.7+20.7\\
% SepCov~\cite{Niklaus_ICCV_2017} & PFRB~\cite{yi2019progressive} & - & - 										& - & - 			& - & - 			& - & - & - & - \\
\hline
DAIN~\cite{DAIN} & Bicubic & 23.64 & 0.6315     & 32.56 & 0.8922 			& 30.82 & 0.8654 			& 29.20 & 0.8309    & 26.61 & 0.7453 & 24.0 \\
DAIN~\cite{DAIN} & RCAN~\cite{zhang2018residual} & 25.07 & 0.7281 										& 35.27 & 0.9242			& 33.82 & 0.9146			& 32.26 & 0.8974 & 26.86 & 0.7489 & 24.0+16.0 \\
DAIN~\cite{DAIN} & RBPN~\cite{haris2019recurrent} & 25.81 & 0.7576 										& 35.67 & 0.9304 			& 34.53 & 0.9262        & 33.01 & 0.9097        & 30.29 & 0.8709 & 24.0+12.7 \\
DAIN~\cite{DAIN} & EDVR~\cite{wang2019edvr} & 26.10 & 0.7828        & 35.81 & 0.9323        & 34.66 & 0.9280      & 33.11 & 0.9119        &~\textcolor{blue}{\underline{30.40}} & 0.8749 & 24.0+20.7\\
% DAIN~\cite{DAIN} & PFRB~\cite{yi2019progressive} & - & - 										& - & - 			& - & - 			& - & - & - & - \\
\hline
AdaCoF~\cite{lee2020adacof} & Bicubic & 23.60 & 0.6287 										& 32.57 & 0.8902 			& 30.78 & 0.8641 			& 29.20 & 0.8308    & 26.59 & 0.7449 & 21.8\\
AdaCoF~\cite{lee2020adacof} & RCAN~\cite{zhang2018residual}  & 24.94 & 0.6807       & 35.08 & 0.9068			& 33.73 & 0.9165 			& 31.93 & 0.8915    & 26.83 & 0.7479 & 21.8+16.0\\
AdaCoF~\cite{lee2020adacof} & RBPN~\cite{haris2019recurrent}  & 25.78 & 0.7546      & 35.55 & 0.9266 			& 34.35 & 0.9236        & 33.08 & 0.9108        &30.39 &~\textcolor{blue}{\underline{0.8751}} & 21.8+12.7 \\
AdaCoF~\cite{lee2020adacof} & EDVR~\cite{wang2019edvr}  & 26.02 & 0.7813    & 35.81 & 0.9294        & 34.56 & 0.9262     & 33.21 & 0.9130   & 30.26 & 0.8709 & 21.8+20.7\\
% AdaCoF~\cite{lee2020adacof} & PFRB~\cite{yi2019progressive} & - & - 										& - & - 			& - & - 			& - & - & - & - \\
\hline
\multicolumn{2}{c|}{STARnet~\cite{haris2020space}} & 26.03 & 0.7818         & 36.04 & 0.9342 			& 34.84 & 0.9298 			& 33.01 & 0.9086        & 29.92 & 0.8589 & 111.6 \\
\multicolumn{2}{c|}{Zooming-SloMo~\cite{xiang2020zooming}} 
&~\textcolor{blue}{\underline{26.31}} &~\textcolor{blue}{\underline{0.7976}}
&~\textcolor{blue}{\underline{36.81}} &~\textcolor{blue}{\underline{0.9415}}
&~\textcolor{blue}{\underline{35.41}} &~\textcolor{blue}{\underline{0.9361}}
&~\textcolor{blue}{\underline{33.36}} &~\textcolor{blue}{\underline{0.9138}}
& 30.34 & 0.8713 &~\textcolor{blue}{\underline{11.1}}
\\
\hline
\multicolumn{2}{c|}{MEGAN} 
&~\textcolor{red}{\textbf{26.57}} &~\textcolor{red}{\textbf{0.8044}}      
&~\textcolor{red}{\textbf{37.18}} &~\textcolor{red}{\textbf{0.9446}}			&~\textcolor{red}{\textbf{35.71}} &~\textcolor{red}{\textbf{0.9389}} 	
&~\textcolor{red}{\textbf{33.62}} &~\textcolor{red}{\textbf{0.9171}}
&~\textcolor{red}{\textbf{30.56}} &~\textcolor{red}{\textbf{0.8753}}
&~\textcolor{red}{\textbf{10.7}}\\
\hline\hline
\end{tabular}
}
\vspace{-3mm}
\caption{Quantitative Evaluation of state-of-the-art methods.~\textcolor{red}{\textbf{Red}} and \textcolor{blue}{\underline{blue}} indicate the best and the second best performance, respectively.}
% \caption{Quantitative Evaluation of state-of-the-art methods. Bold fonts indicate the best performance.}
\label{table:quantitative}
\vspace{-5mm}
\end{table*}

\begin{figure*}[t]
%\tiny
%\small
	%\newlength\fsdurthree
	%\setlength{\fsdurthree}{-1.5mm}
	\scriptsize
	\centering
	\begin{tabular}{cc}
	%\tiny
	%\scriptsize
	%\footnotesize
	%\small
		%\hspace{-0.4cm}
        \begin{adjustbox}{valign=t}
		%\tiny
			\begin{tabular}{c}
				\includegraphics[width=0.39\textwidth]{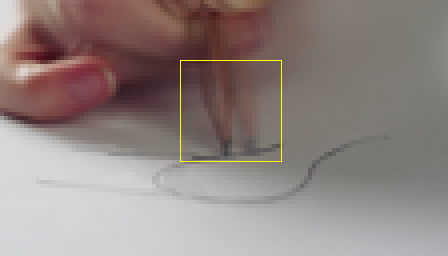}
				\\
				Overlayed LR frames
			
			\end{tabular}
		\end{adjustbox}
		\hspace{-4.3mm}
		\begin{adjustbox}{valign=t}
		%\tiny
			\begin{tabular}{ccccc}
				\includegraphics[width=0.1\textwidth]{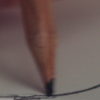} \hspace{-4mm} &
				\includegraphics[width=0.1\textwidth]{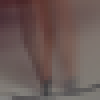} \hspace{-4mm} &
				\includegraphics[width=0.1\textwidth]{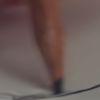} \hspace{-4mm} &
				\includegraphics[width=0.1\textwidth]{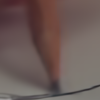}\hspace{-3.5mm} &
				\includegraphics[width=0.1\textwidth]{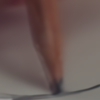}
				\\
				HR \hspace{-4mm} &
    			Overlayed LR \hspace{-4mm} &
				SepConv+EDVR \hspace{-4mm} &
				DAIN+RBPN\hspace{-4mm} &
				AdaCoF+RBPN

				\\
				\includegraphics[width=0.1\textwidth]{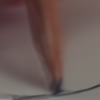} \hspace{-4mm} &
				\includegraphics[width=0.1\textwidth]{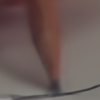} \hspace{-4mm} &
				\includegraphics[width=0.1\textwidth]{Figs/eg1/dainedvr.png} \hspace{-4mm} &
				\includegraphics[width=0.1\textwidth]{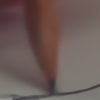}\hspace{-3.5mm} &
				\includegraphics[width=0.1\textwidth]{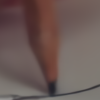}
				\\
				AdaCoF+EDVR \hspace{-4mm} &
				DAIN+EDVR \hspace{-4mm} &
				STARnet\hspace{-4mm} &
				Zooming-SloMo\hspace{-4mm} &
				MEGAN
				\\
			\end{tabular}
			\end{adjustbox}

        %  \\
%         \begin{adjustbox}{valign=t}
% 		%\tiny
% 			\begin{tabular}{c}
% 				\includegraphics[width=0.39\textwidth]{Figs/eg1/lr.png}
% 				\\
% 				Overlayed LR frames
			
% 			\end{tabular}
% 		\end{adjustbox}
% 		\hspace{-4.3mm}
% 		\begin{adjustbox}{valign=t}
% 		%\tiny
% 			\begin{tabular}{ccccc}
% 				\includegraphics[width=0.1\textwidth]{Figs/eg1/hr.png} \hspace{-4mm} &
% 				\includegraphics[width=0.1\textwidth]{Figs/eg1/lr_roi.png} \hspace{-4mm} &
% 				\includegraphics[width=0.1\textwidth]{Figs/eg1/sepedvr.png} \hspace{-4mm} &
% 				\includegraphics[width=0.1\textwidth]{Figs/eg1/dainrbpn.png}\hspace{-3.5mm} &
% 				\includegraphics[width=0.1\textwidth]{Figs/eg1/adarbpn.png}
% 				\\
% 				HR \hspace{-4mm} &
%     			Overlayed LR \hspace{-4mm} &
% 				SepConv+EDVR \hspace{-4mm} &
% 				DAIN+RBPN\hspace{-4mm} &
% 				AdaCoF+RBPN

% 				\\
% 				\includegraphics[width=0.1\textwidth]{Figs/eg1/adaedvr.png} \hspace{-4mm} &
% 				\includegraphics[width=0.1\textwidth]{Figs/eg1/dainedvr.png} \hspace{-4mm} &
% 				\includegraphics[width=0.1\textwidth]{Figs/eg1/dainedvr.png} \hspace{-4mm} &
% 				\includegraphics[width=0.1\textwidth]{Figs/eg1/starnet.png}\hspace{-3.5mm} &
% 				\includegraphics[width=0.1\textwidth]{Figs/eg1/megan.png}
% 				\\
% 				AdaCoF+EDVR \hspace{-4mm} &
% 				DAIN+EDVR \hspace{-4mm} &
% 				STARnet\hspace{-4mm} &
% 				Zooming-SloMo\hspace{-4mm} &
% 				MEGAN
% 				\\
% 			\end{tabular}
% 			\end{adjustbox}
        \\
        \begin{adjustbox}{valign=t}
		%\tiny
			\begin{tabular}{c}
				\includegraphics[width=0.39\textwidth]{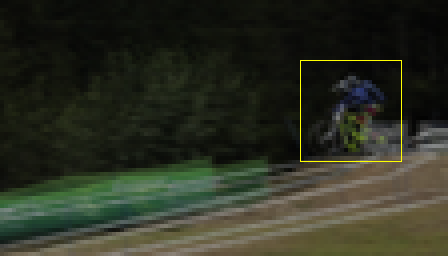}
				\\
				Overlayed LR frames
			
			\end{tabular}
		\end{adjustbox}
		\hspace{-4.3mm}
		\begin{adjustbox}{valign=t}
		%\tiny
			\begin{tabular}{ccccc}
				\includegraphics[width=0.1\textwidth]{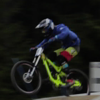} \hspace{-4mm} &
				\includegraphics[width=0.1\textwidth]{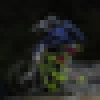} \hspace{-4mm} &
				\includegraphics[width=0.1\textwidth]{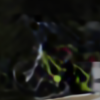} \hspace{-4mm} &
				\includegraphics[width=0.1\textwidth]{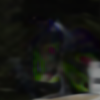}\hspace{-3.5mm} &
				\includegraphics[width=0.1\textwidth]{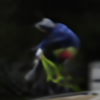}
				\\
				HR \hspace{-4mm} &
    			Overlayed LR \hspace{-4mm} &
				SepConv+EDVR \hspace{-4mm} &
				DAIN+RBPN\hspace{-4mm} &
				AdaCoF+RBPN

				\\
				\includegraphics[width=0.1\textwidth]{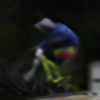} \hspace{-4mm} &
				\includegraphics[width=0.1\textwidth]{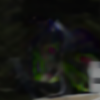} \hspace{-4mm} &
				\includegraphics[width=0.1\textwidth]{Figs/eg2/dainedvr.png} \hspace{-4mm} &
				\includegraphics[width=0.1\textwidth]{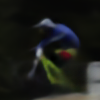}\hspace{-3.5mm} &
				\includegraphics[width=0.1\textwidth]{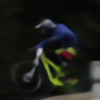}
				\\
				AdaCoF+EDVR \hspace{-4mm} &
				DAIN+EDVR \hspace{-4mm} &
				STARnet\hspace{-4mm} &
				Zooming-SloMo\hspace{-4mm} &
				MEGAN
				\\
			\end{tabular}
			\end{adjustbox}
			
							      \\
         \begin{adjustbox}{valign=t}
		%\tiny
			\begin{tabular}{c}
				\includegraphics[width=0.39\textwidth]{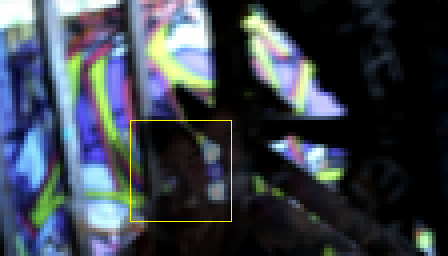}
				\\
				Overlayed LR frames
			
			\end{tabular}
		\end{adjustbox}
		\hspace{-4.3mm}
		\begin{adjustbox}{valign=t}
		%\tiny
			\begin{tabular}{ccccc}
				\includegraphics[width=0.1\textwidth]{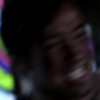} \hspace{-4mm} &
				\includegraphics[width=0.1\textwidth]{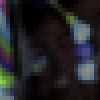} \hspace{-4mm} &
				\includegraphics[width=0.1\textwidth]{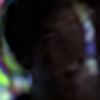} \hspace{-4mm} &
				\includegraphics[width=0.1\textwidth]{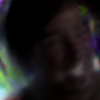}\hspace{-3.5mm} &
				\includegraphics[width=0.1\textwidth]{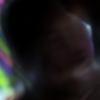}
				\\
				HR \hspace{-4mm} &
    			Overlayed LR \hspace{-4mm} &
				SepConv+EDVR \hspace{-4mm} &
				DAIN+RBPN\hspace{-4mm} &
				AdaCoF+RBPN

				\\
				\includegraphics[width=0.1\textwidth]{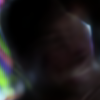} \hspace{-4mm} &
				\includegraphics[width=0.1\textwidth]{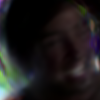} \hspace{-4mm} &
				\includegraphics[width=0.1\textwidth]{Figs/eg3/dainedvr.png} \hspace{-4mm} &
				\includegraphics[width=0.1\textwidth]{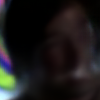}\hspace{-3.5mm} &
				\includegraphics[width=0.1\textwidth]{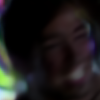}
				\\
				AdaCoF+EDVR \hspace{-4mm} &
				DAIN+EDVR \hspace{-4mm} &
				STARnet\hspace{-4mm} &
				Zooming-SloMo\hspace{-4mm} &
				MEGAN
				% \\
			\end{tabular}
			\end{adjustbox}
	\vspace{-3mm}
	\end{tabular}
    \caption{Visual comparison of Space-time SR results produced by different methods. Our MEGAN achieves more visually pleasant results by handling motion blur and motion aliasing in complex dynamic space-time scenes~(\textbf{Zoomed for visual clarity}).}
    \label{fig:examples}
	\vspace{-5mm}
\end{figure*}

\subsection{Frame Feature Temporal Interpolation}

% We select frame feature temporal interpolation module as our frame feature temporal interpolation network due to its superior performance.
Frame feature temporal interpolation module has been proven to have superior performance in~\cite{xiang2020zooming}, thus we use it in our network to for frame feature interpolation. Given two input feature maps $F^{LR}_{1}$ and $F^{LR}_{3}$, our goal is to learn a feature temporal interpolation function $f(\cdot)$ to estimate the in-between feature map $F^{LR}_{2}$. To accurately capture complex intra-sequence motions~\cite{tian2018tdan}, we adopt the modulated deformable sampling module~\cite{xiang2020zooming} for frame feature sampling. Since the intermediate frame $F^{LR}_{2}$ is not available, we use the deformable convolution to approximate the forward and backward motion conditions between $F^{LR}_{1}$ and $F^{LR}_{3}$. We first use $g_{1}$ to predict the learnable offset~$\Delta{p_{1}}$ for the feature~$F^{LR}_{1}$. With~$\Delta{p_{1}}$ and~$F^{LR}_{1}$, we obtain the sampled features $F^{LR'}_{1}$ by the deformable convolution~\cite{dai2017deformable,zhu2019deformable}. The feature sampling procedure can be defined as follows:
\begin{equation}
    \Delta{p_{1}} = g_{1}(\left[F^{LR}_{1}, F^{LR}_{3}\right]),
    \label{eq: offset1}
\end{equation}
\begin{equation}
    F^{LR'}_{1} = DConv(F^{LR}_{1}, \Delta{p_{1}}),
    \label{eq: sample1}
\end{equation}
where $g_{1}$ denotes the general function including several convolution layers. We follow the same procedure in Equation~\ref{eq: offset1} and~\ref{eq: sample1} to compute the sampled feature~$F^{LR'}_{3}$ from $\Delta{p_{3}}$ and~$F^{LR}_{3}$. Finally, we utilize a simple linear blending function to synthesize the LR feature map. Overall, the general form of the interpolation function is formulated as follows:
\begin{equation}
    F^{LR}_{2} = f(F^{LR}_{1},F^{LR}_{3}) = C_{1}(F^{LR'}_{1}) + C_{3}(F^{LR'}_{3}),
    \label{eq: blend1}
\end{equation}
where $C_{1}$ and $C_{3}$ are $1\times 1$ Conv layer. Similarly, we apply the feature temporal interpolation function $f(\cdot)$ to $\{F^{LR}_{2t-1}\}^{n+1}_{t=1}$. As a result, we can obtain the in-between frame feature maps~$\{F^{LR}_{2t}\}^{n}_{t=1}$.

\subsection{Memory Enhanced Graph Aggregation}
We now have the consecutive frame feature maps~$\{F^{LR}_{t}\}^{2n+1}_{t=1}$ as the input. The module mainly consists of two components: bidirectional deformable ConvLSTM, and long-range memory graph aggregation block (See in Figure~\ref{fig:long-term-memory}). The input is the consecutive frame feature maps~$\{F^{LR}_{t}\}^{2n+1}_{t=1}$. 

% \begin{description}[style=unboxed,leftmargin=0cm]
\paragraph{Deformable ConvLSTM}
We adopt the bidirectional deformable ConvLSTM~\cite{xiang2020zooming} to super-solve video with the complex motion in dynamic scenes. Comparing to naive ConvLSTM, deformable ConvLSTM can better handle lager motion in complex dynamic scenes by aligning hidden and cell states to the reference feature map.
\begin{align}
    \Delta{p^{h}_{t}} &= g^{h}(\left[h_{t-1}, F^{LR}_{t}\right]),\\
    \Delta{p^{c}_{t}} &= g^{c}(\left[c_{t-1}, F^{LR}_{t}\right]),\\
    {h'_{t-1}} &= DConv(h_{t-1},\Delta{p^{h}_{t}}),\\
    {c'_{t-1}} &= DConv(c_{t-1},\Delta{p^{c}_{t}}),\\
    {{h}_{t},{c}_{t}} &= ConvLSTM(h'_{t-1},c'_{t-1},F^{LR}_{t}),
\end{align}
where $\Delta{p^{h}_{t}}$ and $\Delta{p^{c}_{t}}$ refer to the estimated offset of the hidden and cell states. $h'_{t-1}$ and $c'_{t-1}$ are aligned hidden and cell states. $g^{h}$ and $g^{c}$ denote the embedding function which deploys the Pyramid, Cascading and Deformable (PCD) architecture in~\cite{wang2019edvr}.

\paragraph{Long-range Memory Graph Aggregation}
Based on the inter-connected property between space and time domains, we propose a novel long-range memory graph aggregation (LMGA) module for STVSR, drawing inspiration from the context video object detection \cite{chen2020memory}. Different from \cite{chen2020memory}, we utilize Graph Convolutional Network (GCN) to learn appropriate temporal correlations among whole videos. This is because GCN well exploits frame interactions by constructing prior temporal graph for better improving the semantic information among different subsets of frame features. Another major difference is that we use enlarged (interpolated) feature framesets to build our global and local pool, while \cite{chen2020memory} uses Region Proposal Network. Since the enlarged features are often of low quality, our LMGA greatly improves feature quality by using both short and long-range space-time information. In implementations, LMGA includes two operations: frame feature aggregation and graph construction. An example of frame feature aggregation in feature space is shown in Figure~\ref{fig:long-term-memory}. We group the single frame features from the deformable ConvLSTM to form the local pool as $\textbf{L} = \{E^{LR}_{t}\}^{2n+1}_{t=1}$. For long-term global modeling, we randomly shuffle and pick $\tau$ indices from the ordered index sequence $\{1,\ldots,2n+1\}$ to create the global pool~$\textbf{G}=\{E^{G}_{t}\}^{\tau}_{t=1}$. We denote the enriched feature set as~$\textbf{M}=\{E^{M}_t\}_{t=1}^{\varphi=\tau+1}$ by concatenating each local feature and randomly selected global contexts from~$\textbf{G}$. To be specific, the aggregation function can be summarized as:
\vspace{-2mm}
\begin{equation}
    \textbf{M}= N_{g}(\textbf{L},\textbf{G}),
\vspace{-1mm}
\end{equation}
% \vspace{-2mm}
% where $L^{g}=\{L^{g}_t\}_{t=1}^{t=2n+1}$ is the final global-enhanced version of $L$ after the global-local feature aggregation function $N_{g}(\cdot)$ and will be used as inputs for the reconstruction process.
where $N_{g}(\cdot)$ is the global-local feature aggregation function containing three convolutional layers. The spatial-temporal correlations among a collection of frame features can be achieved via a well constructed graph~$\G$, where every frame feature is a vertex and the corresponding edge feature is a similarity-weighted interrelation of two vertices. To exploit the global and local information of space-time domain for STVSR, we propose to perform graph construction in feature domain iteratively.

The graph construction is composed of $K$ weighted graph $\{\G^{k}(\V,\E)\}_{k=1}^{K}$. The graph vertices $\V$ are the enriched feature set~$\textbf{M}$ with the~$\varphi$ nodes, and the correlation (edges) set $\E$ is with the size $|\E|=|\V|\times(\varphi-1)$. Following the message-passing algorithms~\cite{kearnes2016molecular,gilmer2017neural}, we define the learned edge feature~${A}_{p,q}^{k}$ as the non-linear combination of the absolute difference between any two node features ($p$ and $q$), \eg,
% \begin{equation}
% \vspace{-6pt}
% \label{edgefeats}
${A}_{p,q}^{k} = \phi_{{\theta}}(|{E}_{p}^{k} - {E}_{q}^{k}|)$,
% \end{equation}
where $\phi$ is a transformation function (\eg a neural network) with learnable parameters $\theta$. $E_{p}^{k}$ is the feature embedding for the node $p$ of $\G^{k}$. In this work, we use stacked convolutional layers to embed the node features to get the weighted adjacency ${A}_{p,q}^{k}$ between nodes $p$ and $q$. The adjacent matrix is normalized to a stochastic matrix by using a softmax along each row. 

Inspired by GCN~\cite{garcia2018fewshot}, we aggregate $\varphi$ node embeddings with its neighbours using~\textit{adjacency operator} $A:{E}\mapsto A({E})$ where $A(E_{p}):= \sum_{q\sim p }A_{p,q} {{E}_q}$, with $p\sim q~\text{if and only if}~(p,q)\in\E$. Let us denote ${E}^{k}_{p,r}$ as the $r$-th channel of the embedding of node $p$ in $\G^{k}$. Consequently, the $k$-th GCN layer $\text{Gc}(\cdot)$ is conducted as:
\begin{equation}
\label{gnneq}
{E}^{k+1}_{p,r'} = \text{Gc}({E}^{k}) = \rho \left( \sum_{r=1}^{r_k} \sum_{\upsilon \in \N_{p}} A_{p,\upsilon} {E}_{p,r}^{k}\ast \theta_{r',r}^{k} \right)~\forall r',p 
\end{equation}
where $\theta\in\R^{r_{k}\times r_{k+1}\times b \times b}$ are learned parameters of the non-linear transformation.~$r_k$, $r_{k+1}$, and $b$ denotes the size of the input features and the output features, 2D convolutional kernel size, respectively.~$\N(p)$ is the neighbour node set of the node $p$.$\ast$ denotes the convolution operation. $\rho$ is the operation of Leaky ReLU~\cite{xu2015empirical}. To refine the enriched intermediate features set~$\{E_{t}^{K+1}\}_{t=1}^{(2n+1)}$, we use a small network containing two convolutional layers for embedding features, and then concatenate them to obtain $\{E'_{{t}^{K+1}}\}_{{t=1}^{2n+1}}$. The refined features appears to consistently enable the network with more expressive power, which provides a better starting point for training a model for downstream tasks. As a result, we use LMGA-refined features $F^{M}=\{E_{t}^{K+1'}\}_{t=1}^{2n+1}$ as the input for the subsequent layers of the network. Please see Figure~\ref{fig:long-term-memory} for more details.

\subsection{Progressive Fusion Reconstruction}
In the reconstruction process $\Psi$, we first utilize $m_{2}$ PFRDBs \cite{yi2019progressive} to progressively learn intra-frame spatial representations and inter-frame temporal correlations. After that, we concatenate $m_{3}$ residual blocks \cite{lim2017enhanced} to further enable very deep network for video super resolution tasks. For up-sampling, we perform $\times 2$ upsample \textit{PixelShuffle} operations~\cite{shi2016real,lim2017enhanced}. The last Conv layer fuses all the feature maps to reconstruct HR video sequence~${I}^{HR}\!=\!\{I^{HR}_{t}\}^{2n+1}_{t=1}$. We obtain the final HR output via the reconstruction module, \eg,~${I}^{HR}=\Psi(F^{M})$.

% In the reconstruction process~$\Psi$, we first utilize the non-local block~\cite{wang2018non} to encourage channel-wise features to better capture different types of information. We then add~$m_{2}$ PFRDBs~\cite{yi2019progressive} to progressively learn intra-frame spatial representations and inter-frame temporal correlations. After that, we concatenate $m_{3}$ residual blocks~\cite{lim2017enhanced} to further enable very deep network for video super resolution tasks. For up-sampling, we perform $\times 2$ upsample~\textit{PixelShuffle} operations~\cite{shi2016real,lim2017enhanced}. The last Conv layer fuses all the feature maps to reconstruct HR video sequence~${I}^{HR} = \{I^{HR}_{t}\}^{2n+1}_{t=1}$. We obtain the final HR output via the reconstruction module, \eg,~${I}^{HR}=\Psi(F^{M})$.

\begin{figure*}[t]
%\tiny
%\small
	%\newlength\fsdurthree
	%\setlength{\fsdurthree}{-1.5mm}
	\scriptsize
	\centering
	\begin{tabular}{cc}
	%\tiny
	%\scriptsize
	%\footnotesize
	%\small
		%\hspace{-0.4cm}
        \begin{adjustbox}{valign=t}
		%\tiny
			\begin{tabular}{c}
				\includegraphics[width=0.33\textwidth]{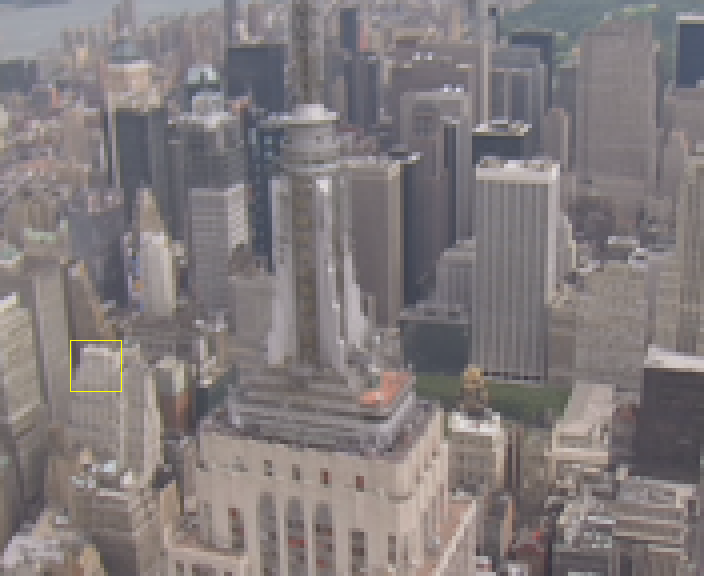}
				\\
				Overlayed LR frames
			
			\end{tabular}
		\end{adjustbox}
		\hspace{-4.3mm}
		\begin{adjustbox}{valign=t}
		%\tiny
			\begin{tabular}{ccccc}
				\includegraphics[width=\widthscale \textwidth]{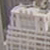} \hspace{-4mm} &
				\includegraphics[width=\widthscale \textwidth]{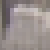} \hspace{-4mm} &
				\includegraphics[width=\widthscale \textwidth]{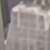} \hspace{-4mm} &
				\includegraphics[width=\widthscale \textwidth]{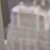}\hspace{-3.5mm} &
				\includegraphics[width=\widthscale \textwidth]{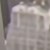}
				\\
				HR \hspace{-4mm} &
    			Overlayed LR \hspace{-4mm} &
				SepConv+EDVR \hspace{-4mm} &
				DAIN+RBPN\hspace{-4mm} &
				AdaCoF+RBPN

				\\
				\includegraphics[width=\widthscale \textwidth]{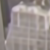} \hspace{-4mm} &
				\includegraphics[width=\widthscale \textwidth]{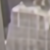} \hspace{-4mm} &
				\includegraphics[width=\widthscale \textwidth]{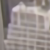} \hspace{-4mm} &
				\includegraphics[width=\widthscale \textwidth]{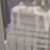}\hspace{-3.5mm} &
				\includegraphics[width=\widthscale \textwidth]{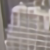}
				\\
				AdaCoF+EDVR \hspace{-4mm} &
				DAIN+EDVR \hspace{-4mm} &
				STARnet\hspace{-4mm} &
				Zooming-SloMo\hspace{-4mm} &
				MEGAN
				% \\
			\end{tabular}
			\end{adjustbox}
	\vspace{-3mm}
	\end{tabular}
	\caption{Visual comparison. Our MEGAN recovers more detailed textures with sharper edges (\textbf{Zoomed for visual clarity}).}
	\label{fig:vidresult}
	\vspace{-5mm}
\end{figure*}

% \vspace{-2mm}
\subsection{Loss Function}
In the training process, our proposed network is optimized by the following objective function in an end-to-end manner:
\vspace{-3mm}
\begin{equation}
    L_{\mathrm{MEGAN}} = \sqrt{\|{I}^{GT} - {I}^{HR}\|^{2} + \epsilon^{2}},
\end{equation}
where ${I}^{GT}$ denotes the ground-truth HR video sequence. We adopt Charbonnier penalty function~\cite{lai2017deep} as the loss term and empirically set $\epsilon = 1e^{-3}$.

% we utilize $m_{1} = 3$ Conv layers, $m_{2} = 40$ PFRBs,  and $m_{3} = 20$ residual blocks in the residual non-local feature extractor, and progressive fusion reconstruction network, respectively. 
\subsection{Implementation Details}
In the proposed MEGAN, we utilize $m_{1} = 3$ Conv layers in the residual non-local attention module. For the progressive fusion reconstruction network, we set~$m_{2}, m_{3}=5, 20$. In the LMGA, we set $\tau$ as $4$, and $k$ as $2$. During the training, we select the odd-indexed $4$ frames and the corresponding $7$ consecutive frames as the LR input and ground-truth HR, respectively. We augment the training data in these ways, such as rotation, flipping, and randomly reversing the direction of video sequences. On the image level, each video frame is down-sampled by $4$, and a random crop of $32\times 32$ on LR images is performed. We train our network on $2$ NVIDIA 1080Ti GPUs using the Adam optimizer~\cite{kingma2014adam}. Using a cosine annealing strategy~\cite{loshchilov2016sgdr}, the initial learning rate starts with $1e^{-4}$, and is annealed to~$1e^{-7}$ after $20000$ iterations. In the training process, we set batch size as $16$. For fair comparisons, we evaluate all the methods on the same single Nvida 1080Ti GPU. The number of parameters of MEGAN is $10.7$ Million.

% All code will be released after publication.

% 10.7

% $N_{g}(\cdot)$ is the small embedding network containing three convolutional layers. All code will be released after publication.

% The number of parameters of MEGAN is $15.59$~Million. All code will be released after publication.
% Please follow the steps outlined below when submitting your manuscript to
% the IEEE Computer Society Press.  This style guide now has several
% important modifications (for example, you are no longer warned against the
% use of sticky tape to attach your artwork to the paper), so all authors
% should read this new version.

%-------------------------------------------------------------------------
\section{Experiments}
\paragraph{Experimental Setup}
In this study, we utilize the Vimeo-90K dataset~\cite{xue2019video} for training to demonstrate the fidelity and robustness of our proposed method. The dataset consists of over 60,000 7-frame training video sequences with the resolution of $256\times 448$. The resolution of down-scaled LR frames is $64\times 112$. To evaluate the performance of different methods, we select several datasets, including Vid4~\cite{liu2011bayesian}, Vimeo~\cite{xue2019video}, and Adobe240~\cite{su2017deep} testsets. Following the setting in~\cite{RBPN2019}, we split Vimeo testset into fast motion, medium motion, and slow motion subsets. For evaluation, we utilize two widely-used image quality metrics to quantitatively validate the STVSR performance, including peak signal-to-noise ratio (PSNR) and structural similarity index metrics~(SSIM)~\cite{wang2004image}.

\paragraph{Comparison to State-of-the-art}
In this study, we compare the proposed MEGAN with the state-of-the-art methods. For clarity, we categorize the methods into one-stage and two-stage. For one-stage methods, we choose Zooming SlowMo~\cite{xiang2020zooming} and STARnet~\cite{haris2020space} for comparison. For two-stage methods, we pick SepConv~\cite{Niklaus_ICCV_2017}, Super-SloMo~\cite{jiang2018super}, DAIN~\cite{DAIN}, AdaCoF~\cite{lee2020adacof} as Time SR and RCAN~\cite{zhang2018residual}, RBPN~\cite{haris2019recurrent}, EDVR~\cite{wang2019edvr}, PFRB~\cite{yi2019progressive} as Space SR.

We evaluate the proposed method against the state-of-the-art methods on the Vid4 and Vimeo datasets. The typical results are given in Figures~\ref{fig:examples} and~\ref{fig:vidresult}. To further evaluate the robustness of our method, the recovered features are zoomed in Figures~\ref{fig:examples} and~\ref{fig:vidresult}. We can clearly observe that our MEGAN provides more visually pleasant results with finer details and sharper edges than the competing methods. The quantitative results are summarized in Table~\ref{table:quantitative}. We can see that our model obtains $0.26$, $0.37$, $0.30$, $0.26$, and~$0.22$~dB PSNR gains over the best one-stage-based model Zooming-SloMo on Vid4, Vimeo-Fast, Vimeo-Medium, Vimeo-Slow, and Adobe240, and also outperforms the best two-stage method DAIN+EDVR by $0.47$,$1.37$,$1.05$,$0.51$ and $0.16$~dB on Vid4, Vimeo-Fast, Vimeo-Medium, Vimeo-Slow, and Adobe240 in terms of PSNR, respectively. To better illustrate the robustness for space-time upsampling, we also compare our proposed method with 18 state-of-the-art methods on Adobe240~\cite{liu2011bayesian} dataset. As shown in Table~\ref{table:quantitative}, the proposed MEGAN consistently outperforms all the evaluated methods. This suggests that the proposed MEGAN is capable of recovering better results with finer details by capturing long-term spatial-temporal information. These comparisons intuitively show that it is essential to access more global and local information from the space-time domain to handle motion blur and motion aliasing in complex dynamic space-time scenes. In other words, the results demonstrate that the proposed MEGAN reconstructs more detailed results with more sharper contexts, as shown in Figures~\ref{fig:examples} and~\ref{fig:vidresult}.
\vspace{-5pt}
% by obtaining more detailed information in the feature domain.

% Zooming-SloMo outperforms the best two-stage method DAIN+EDVR by $0.21$~dB on Vid4. It is not surprising that the one-stage-based methods obtain favorable PSNR and SSIM values since the one-stage-based methods can capture the correlations between the spatial and temporal dimensions to increase the spatial-temporal resolution. In addition, we can see that our model obtains $0.17$, $0.33$, $24$, and~$0.22$~dB PSNR gains over the best one-stage-based model Zooming-SloMo on Vid4, Vimeo-Fast, Vimeo-Medium and Vimeo-Slow, and also outperforms the best two-stage method DAIN+EDVR by $0.38$~dB on Vid4, $1.30$~dB on Vimeo-Fast, $0.99$~dB on Vimeo-Medium, and $0.47$~dB on Vimeo-Slow in terms of PSNR. These comparisons show that it is essential to access more global and local information from the space-time domain to handle motion blur and motion aliasing in complex dynamic space-time scenes. In other words, the results demonstrate that the proposed MEGAN reconstructs more pleasant results with more detailed contexts.

\begin{figure}[t]
    % \footnotesize
    \centering
    \begin{subfigure}[b]{0.48\columnwidth}
    \centering
    \includegraphics[width=\columnwidth]{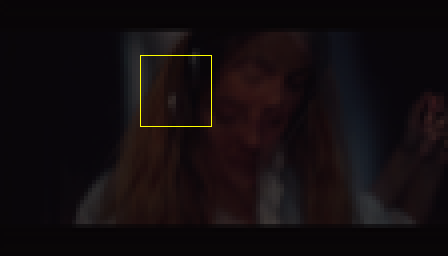}
    \caption*{Overlayed LR}
     \end{subfigure}
    \begin{subfigure}[b]{0.48\columnwidth}
    \centering
    \includegraphics[width=\columnwidth]{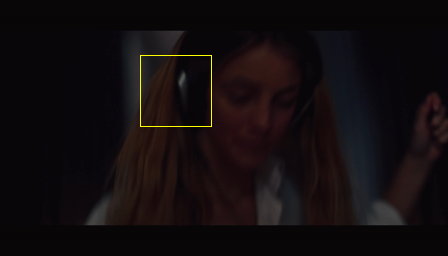}
    \caption*{HR}
    \end{subfigure}     
    % \hfill
    \par\vfill

    \begin{subfigure}[b]{0.48\columnwidth}
    \centering
    \includegraphics[width=\columnwidth]{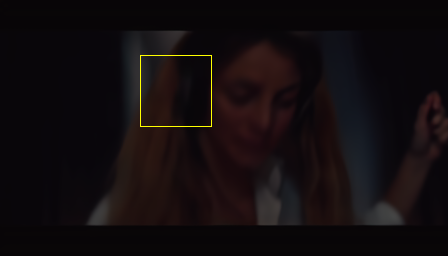}
    \caption*{w/o LMGA}
    \end{subfigure}
    \begin{subfigure}[b]{0.48\columnwidth}
    \centering
    \includegraphics[width=\columnwidth]{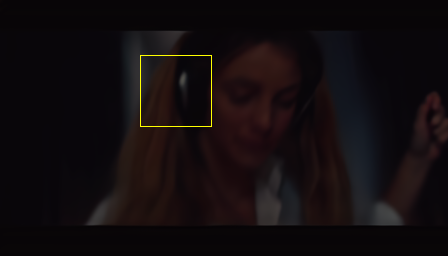}
    \caption*{w/ LMGA}
    % \label{fig:my_label}
    \end{subfigure}
    \vspace{-3mm}
    \caption{Illustration of effectiveness of LMGA. As shown in the yellow boxes, we can observe that w/ LMGA module recovers more richer details (\eg, headphone)}
    \label{fig:LMGA}
    % \vspace{-3mm}
\end{figure}

\begin{table}[t]
\centering
\resizebox{0.8\linewidth}{!}{
\begin{tabular}{c c c c c}
\hline\hline
Method & \multicolumn{2}{c}{Vid4} & \multicolumn{2}{c}{Vimeo-Fast} \\
& PSNR & SSIM & PSNR & SSIM \\
\hline\hline
w/o LMGA & 26.40 & 0.8001                                        & 36.89 & 0.9419            \\
w/o NLRB & 26.44 & 0.8017 										& 37.01 & 0.9430            \\
% \hline
Ours &~\textbf{26.57} &~\textbf{0.8044}         &~\textbf{37.18} &~\textbf{0.9446}		 \\
\hline\hline
\end{tabular}
}
\vspace{-3mm}
\caption{Ablation study on LMGA and NLRB over Vid4 and Vimeo-Fast dataset.}
% Bold fonts indicate the best performance.}
\label{table:effectiveness}
% \vspace{-3mm}
\end{table}

\section{Ablation Study}
We conduct a thorough ablation analysis to demonstrate the effectiveness of different key components of MEGAN. Detailed results are found in the supplementary material.

% The results are reported in Table~\ref{table:effectiveness}.

\paragraph{Effectiveness of LMGA}
% To evaluate the proposed LMGA block, we employ our proposed method w/ and w/o LMGA block as the baseline. Table~\ref{table:effectiveness} shows that our baseline network with LMGA block yield inspiring increase in space-time SR in term of PRSN and SSIM value. The visual comparisons are depicted in Figure~\ref{fig:LMGA}. These SR results results that wit LMGA block could further empower the network to preserve sharp spatial features by combining information in the temporal direction from fast dynamic events.
We first explore the effectiveness of the proposed LMGA block as it is the key component in our model. We establish our baseline network (w/o LMGA model) by removing the LMGA block from our model. The quantitative results are provided in Table~\ref{table:effectiveness}. The increased performance in terms of PSNR value over the the Vimeo-Fast dataset, from $36.89$ to~$37.18$ dB, demonstrates that the LMGA module can better capture fast motions by aggregating global temporal information. The visual comparisons are depicted in Figure~\ref{fig:LMGA}. From the visual results, we can see that the LMGA block enables the model to accurately reconstruct frames with more details. In particular, the ``w/o LMGA'' lower bound fails to realize the headphone on the woman's head, but our model provides better visualization of the headphone. These observations demonstrate that adding the LMGA block to better utilize the global and local temporal contexts can further encourage our model to obtain more accurate spatial-temporal representations from fast dynamic video sequences.

\begin{figure}
    \centering
    \begin{subfigure}[b]{0.48\columnwidth}
    \centering
    \includegraphics[width=\columnwidth]{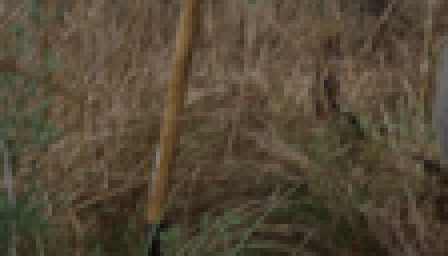}
    \caption*{Overlayed LR}
     \end{subfigure}
    \begin{subfigure}[b]{0.48\columnwidth}
    \centering
    \includegraphics[width=\columnwidth]{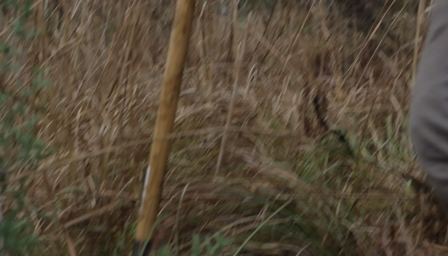}
    \caption*{HR}
    \end{subfigure}     
    % \hfill
    \par\vfill

    \begin{subfigure}[b]{0.48\columnwidth}
    \centering
    \includegraphics[width=\columnwidth]{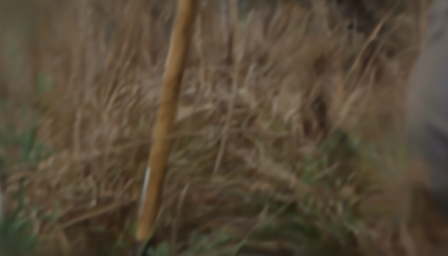}
    \caption*{w/o NLRB}
    \end{subfigure}
    \begin{subfigure}[b]{0.48\columnwidth}
    \centering
    \includegraphics[width=\columnwidth]{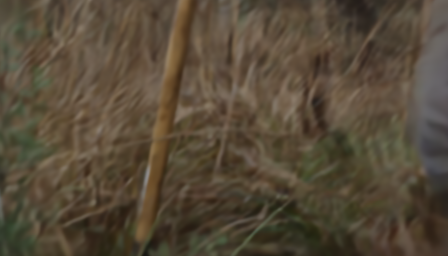}
    \caption*{w/ NLRB}
    % \label{fig:my_label}
    \end{subfigure}
    \vspace{-3mm}
    \caption{Illustration of robustness of LNRB. As we see, w/ NLRB is able to contain richer and sharper details from the overlayed LR image especially in the regions of surrounding areas.}
    \label{fig:NLRB}
    \vspace{-5mm}
\end{figure}
% `` xxxx ''

\paragraph{Effectiveness of NLRB}
% To investigate the effect of the NLRB block, our baseline network uses the proposed method w/o NLRB block, i.e., removing two NLRB block when training the network. As shown in Table~\ref{table:effectiveness}, deep non-local residual learning strategy can significantly increase resolution both in time and space domain. Its support is further illustrated in Figure~\ref{fig:NLRB}. We can clearly see that it can significantly alleviate motion blur and recover very fast dynamic events sequence with sharp textures. preserves better textural details and 
% We remove two NLRB blocks from the proposed method to create a w/o NLRB model and use it as the baseline. 
We then investigate the effectiveness of the NLRB block. Our baseline network adopts the proposed network w/o NLRB block, i.e., removing two NLRB blocks when training the network. The results in Table~\ref{table:effectiveness} show that the deep non-local residual learning strategy helps improve the performances on two benchmark datasets. In Figure~\ref{fig:NLRB}, it is observed that our model preserves better textural details of grasses~(shown at left), and captures the human leg motion~(shown at right). In comparison, the ``w/o NLRB'' model has limited ability to recover fine details of grasses in the central area, but the surrounding area is still blurry. These results show that the NLRB block enables the model to capture long-range dependencies in the spatial dimension, while the channel-wise attention enables the model to attend different types of feature representations. NLRB can further encourage the network to retain high-frequency details by incorporating more spatial information. Its support is further illustrated in Figure~\ref{fig:NLRB}. We can see that NLRB is capable of retraining low- and high-frequency information to reconstruct more realistic video sequences. These observations demonstrate the effectiveness of NLRB in providing more detailed textures.
% can significantly alleviate motion blur and recover very fast dynamic event sequences with sharp textures.pushes the model to achieve better SR performance on Vid4 and Vimeo-Fast datasets. 
\section{Conclusion}
In this work, we present a unified framework for space-time video super-resolution. Aided by temporal and spatial information, our approach learns complex spatial-temporal features to reconstruct a video sequence of high space-time resolution. To this end, we propose a novel long-range memory graph aggregation (LMGA) module to enhance the features of key frames with global and local information in the dynamic space-time scene. We also introduce the deep non-local residual learning to enable our model to adaptively learn mixed attention hierarchical features from multiple video sequences. Furthermore, we adopt the progressive fusion reconstruction network to capture long-range spatial-temporal dependencies. In general, the proposed MEGAN has achieved the superior performance among all the involved methods in both static and dynamic scenes.

{\small
\bibliographystyle{ieee_fullname}
\bibliography{egbib}
}

\end{document}